\documentclass[sigconf]{acmart}
\AtBeginDocument{%
  \providecommand\BibTeX{{%
    \normalfont B\kern-0.5em{\scshape i\kern-0.25em b}\kern-0.8em\TeX}}}

\copyrightyear{2023}
\acmYear{2023}
\setcopyright{acmlicensed}
\acmConference[SIGIR '23]{Proceedings of the 46th International ACM SIGIR Conference on Research and Development in Information Retrieval}{July 23--27, 2023}{Taipei, Taiwan}
\acmBooktitle{Proceedings of the 46th International ACM SIGIR Conference on Research and Development in Information Retrieval (SIGIR '23), July 23--27, 2023, Taipei, Taiwan}
\acmPrice{15.00}
\acmDOI{10.1145/3539618.3591721}
\acmISBN{978-1-4503-9408-6/23/07}
\acmSubmissionID{3080}

\usepackage{graphicx}
\usepackage{amsmath}
\usepackage{amssymb}
\usepackage{booktabs}
\usepackage{wrapfig}
\usepackage{multirow}
\usepackage{stfloats}
\usepackage{float}
\usepackage[capitalize]{cleveref}
\crefname{section}{Sec.}{Secs.}
\Crefname{section}{Section}{Sections}
\Crefname{table}{Table}{Tables}
\crefname{table}{Tab.}{Tabs.}

\newcommand{\ie }{\emph{i.e.}}
\newcommand{\eg}{\emph{e.g.}}
\newcommand{\etc}{etc}

\begin{document}

\title{MAMO: Fine-Grained Vision-Language Representations Learning with Masked Multimodal Modeling}

\author{Zijia Zhao$^{1,3*\dagger}$,~~~Longteng Guo$^{2*}$,~~~Xingjian He$^1$,~~~Shuai Shao$^2$,~~~Zehuan Yuan$^2$,~~~Jing Liu$^{1,3\ddagger}$}\thanks{*Equal contribution.}\thanks{$^\dagger$This work was performed while Zijia worked as an intern at ByteDance.}\thanks{$^\ddagger$Corresponding author.}

\affiliation{
 \institution{$^1$Laboratory of Cognition and Decision Intelligence for Complex Systems, \\Institute of Automation, Chinese Academy of Sciences}
  \country{}
}

\affiliation{
 \institution{$^2$Bytedance Inc.~~~$^3$School of Artificial Intelligence, University of Chinese Academy of Sciences}
  \country{}
}

\affiliation{ 
\tt\small zhaozijia2021@ia.ac.cn, \{xingjian.he,jliu\}@nlpr.ia.ac.cn, \tt\small \{guolongteng.lt,shaoshuai.0516,yuanzehuan\}@bytedance.com   \country{}}
\renewcommand{\authors}{Zijia Zhao, Longteng Guo, Xingjian He, Shuai Shao, Zehuan Yuan, Jing Liu}
\renewcommand{\shortauthors}{Zijia Zhao, Longteng Guo, Xingjian He, Shuai Shao, Zehuan Yuan, Jing Liu}

\begin{abstract}
Multimodal representation learning has shown promising improvements on various vision-language tasks (\eg, image-text retrieval, visual question answering, \etc) and has significantly advanced the development of multimedia information systems. Most existing methods excel at building global-level alignment between vision and language while lacking effective fine-grained image-text interaction. In this paper, we propose a jointly masked multimodal modeling method to learn fine-grained multimodal representations. Our method performs joint masking on image-text input and integrates both implicit and explicit targets for the masked signals to recover. The implicit target provides a unified and debiased objective for vision and language, where the model predicts latent multimodal representations of the unmasked input. The explicit target further enriches the multimodal representations by recovering high-level and semantically meaningful information: momentum visual features of image patches and concepts of word tokens. Through such a masked modeling process, our model not only learns fine-grained multimodal interaction, but also avoids the semantic gap between high-level representations and low- or mid-level prediction targets (\eg , image pixels, discrete vision tokens), thus producing semantically rich multimodal representations that perform well on both zero-shot and fine-tuned settings. Our pre-trained model (named MAMO) achieves state-of-the-art performance on various downstream vision-language tasks, including image-text retrieval, visual question answering, visual reasoning, and weakly-supervised visual grounding.
\end{abstract}






\maketitle
\section{Introduction}
\label{sec:intro}
Vision-Language Pre-training (VLP) is an emerging research topic in multimedia information systems. It aims to learn the interaction between image and text and produce semantically rich multimodal representations that transfer well to downstream Vision-and-Language (V+L) tasks including image-text retrieval, visual question answering, \etc. In order to learn cross-modal interaction, most existing VLP methods~\cite{clip,align,albef,uniter,vlbert,oscar} rely on the consistency between the global views of image and text, using pre-training objectives like image-text contrastive loss~\cite{albef} and image-text matching loss~\cite{vlbert}. While effective, such global interaction fails to model the subtle local association within image-text pairs. The fine-grained interaction between image patches and word tokens is therefore lacking.

\begin{figure}[t]
\centering
\includegraphics[width=\linewidth]{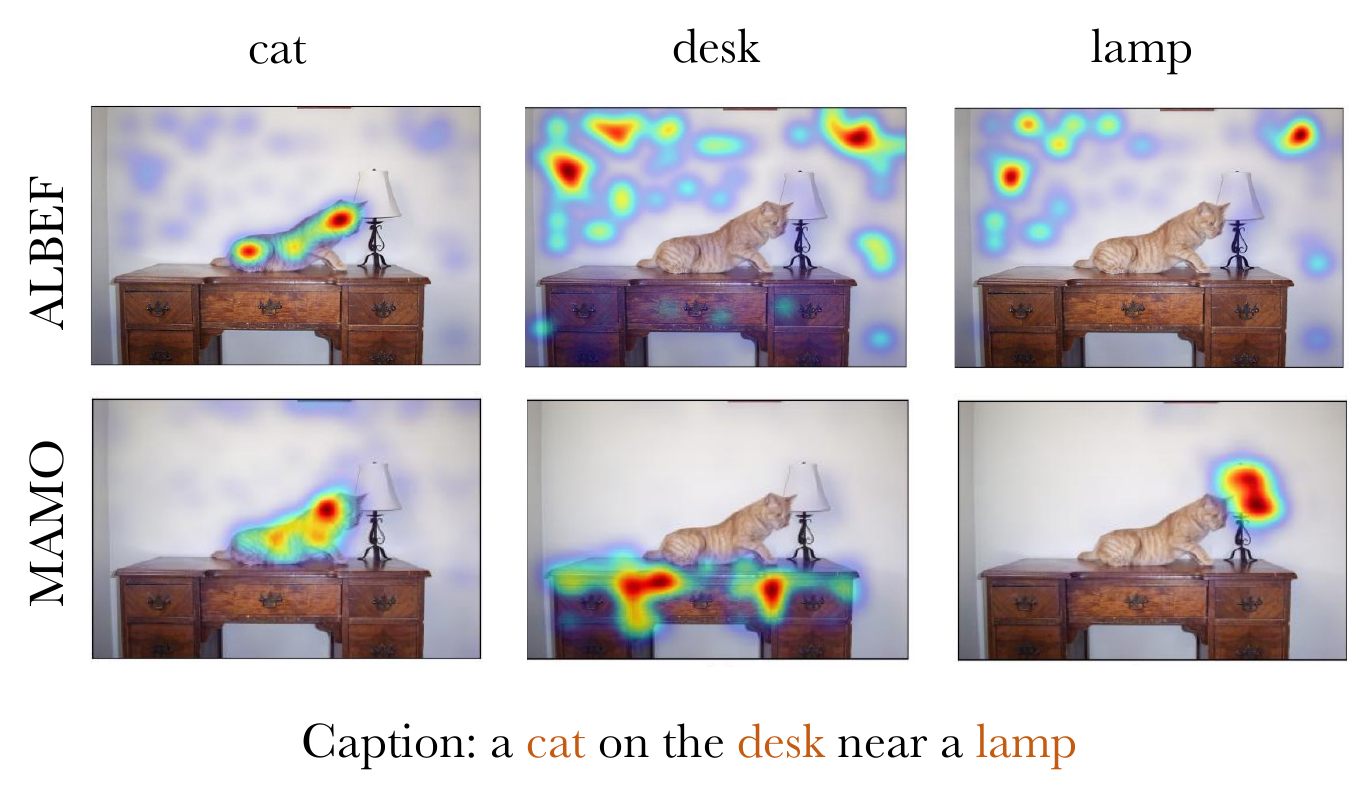}
\caption{Per-word Grad-CAM visualization of word to image attention. Comparing with ALBEF~\cite{albef}, MAMO can focus on corresponding region for each word more precisely, indicating more fine-grained interactions between word and image patches are built. }
\label{fig:intro_case}
\end{figure}

Masked signal modeling is an effective self-supervised pre-training task that masks a portion of input signals and tries to predict these masked signals from the visible ones. It has been actively explored in natural language processing (NLP) and computer vision (CV) separately, and has brought powerful generalization performance across diverse downstream tasks. For example, BERT~\cite{bert} formulates masked language modeling (MLM) to predict masked linguistic tokens, while MAE~\cite{mae} and BEiT~\cite{beit} formulate masked image modeling (MIM) to reconstruct raw pixels and dVAE visual tokens~\cite{dalle} of image patches respectively. 
In the domain of VLP, however, there is a lack of a jointly masked signal modeling method for both vision and language modalities. Although previous VLP methods ~\cite{uniter,albef} have adopted conditional MLM to predict masked words given \textit{unmasked} image and other words, however, masking of the image side has not been fully explored. As a result, the images’ internal structures and their interactions with text tokens are not sufficiently learned, as is shown in~\cref{fig:intro_case}.

The challenge of designing a jointly masked signal modeling method for VLP lies in the natural differences between image and text modalities --- image is continuous, low-level, highly redundant raw signals, while text tokens are discrete, high-level, highly compressed concepts generated by humans. This phenomenon raises two questions: (1) How to design a unified prediction target that applies to masked multimodal data composing both continuous visual signals and discrete text tokens? (2) How to avoid the semantic gap between the learning of high-level representations and the prediction of low-level image signals?

In this paper, we propose MAsked Multimodal mOdel (MAMO), a VLP model with a jointly masked learning strategy on both vision and language modalities. 
MAMO performs joint masking on image-text input and integrates both implicit and explicit targets for the masked signals to recover. The \textit{implicit} target provides a unified and debiased objective for vision and language, the core idea of which is to predict latent multimodal representations of the masked signals under the guidance of a self-distillation teacher that takes the unmasked view as input. 
Such a bootstrapping latent target avoids learning biases in modality-specific designs.
While the implicit prediction target performs effective empirically, it can collapse into meaningless solutions, \eg, outputting the same vector for all tokens. To further enrich the multimodal representations and avoid such potential trivial solutions, we also add auxiliary \textit{explicit} prediction targets that are naturally distinguishable on each masked position. These targets are explicit in that they are semantically meaningful features or concepts extracted from the raw data. Particularly, for the masked image tokens, instead of reconstructing low-level raw pixels~\cite{mae} or predicting mid-level pre-defined visual tokens~\cite{beit} (encapsulating mostly patch details, \eg color and texture, according to~\cite{ibot}), they are enforced to predict the high-level momentum visual features extracted from the image encoder. As for masked text tokens, we directly predict word tokens since they are already high-level concepts.

MAMO naturally addresses the aforementioned two questions: First, the prediction targets of both masked vision and language signals are unified as regressing the implicit latent multimodal representations; Second, our implicit and explicit prediction targets are both high-level representations and features/concepts, thus avoiding the semantic-gap caused by predicting low- or mid-level image signals. Such a masked modeling process enforces the model to build both intra- and inter-modality interactions between image patches and word tokens, and produce fine-grained and semantically rich multimodal representations that perform well on both zero-shot and fine-tuned settings.

We demonstrate the effectiveness of MAMO on various downstream V+L tasks including image-text retrieval, visual question answering, visual reasoning, and weakly-supervised visual grounding. MAMO achieves substantial improvements over existing state-of-the-art methods. 
On zero-shot image-text retrieval, MAMO even outperforms methods that are
pre-trained on orders of magnitude larger datasets, \eg, it outperforms the state-of-the-art methods, CLIP~\cite{clip} and ALIGN~~\cite{align}, by absolute improvements of $11.9\%$ and $4.9\%$ on image retrieval R@1, respectively.  On fine-tuned image-text retrieval, MAMO also outperforms other methods (\eg, ALBEF~\cite{albef}, METER~\cite{meter}, VLMo~\cite{vlmo}) with a large margin. Quantitative and qualitative analyses on MAMO using Grad-CAM~\cite{gradcam} further demonstrate its ability to perform more fine-grained vision-language interaction.
Our contributions are summarized as follows:
\begin{itemize}
\item We propose a jointly masked multimodal modeling method that integrates both implicit and explicit prediction targets to learn fine-grained multimodal representations.
\item Our implicit target provides a unified and debiased objective for VLP, and our explicit target shows that high-level momentum visual features can serve as a better auxiliary target for masked images compared with low-level pixels or mid-level visual tokens.
\item Qualitative and quantitative results across a broad range of downstream tasks show that our method learns fine-grained and transferable vision-language representations.
\end{itemize}

\section{Related Work}
\subsection{Vision-Language Representation Learning}

Depending on how vision and language modalities interact, most previous VLP methods fall into two categories: shallow interaction and deep interaction. Shallow interaction methods~\cite{clip,align} use light-weight operations (\eg, dot product) for interaction while deep interaction methods~\cite{vlbert, uniter, lxmert, visualbert, oscar, vinvl} use deep networks (\eg, a transformer-based multimodal encoder) to fuse image and text features. ALBEF~\cite{albef} combines the above two types of methods, learning both shallow and deep interactions in a single framework. To train such interaction networks, previous VLP methods often employ contrastive learning~\cite{albef} or image-text matching~\cite{visualbert} as pre-training tasks, which excel at learning global-level image-text alignment, but lack effective modeling of fine-grained interaction between image patches and word tokens. 
MAMO inherits the architecture of ALBEF to combine shallow and deep interactions and further introduces a new masked multimodal modeling method to enforce fine-grained multimodal interaction. 

\subsection{Masked Signal Modeling}
Masked signal modeling has been actively explored in NLP and CV separately. The prediction target for masked signals to recover is one of the main differences in previous works. In NLP, word token is the most commonly used prediction target. In CV, various targets are explored, \eg, raw pixel~\cite{mae,simmim}, HOG features~\cite{maskfeat}, visual tokens~\cite{beit} from a pre-trained dVAE~\cite{dalle}, visual tokens from a momentum model~\cite{ibot}, \etc.

In VLP, conditional MLM is a commonly used pre-training objective, which predicts masked text tokens based on unmasked images. 
Some works perform MIM in VLP.
UNITER~\cite{uniter} applies masking on pre-extracted regional features and let the model predict the class label or feature of those regions.
OSCAR~\cite{oscar} inputs additional object tags into the model and adds a mask-and-predict task on these object tags.
However, these methods require a pre-trained model, \eg, object detector like Faster R-CNN~\cite{fasterrcnn}, to extract the prediction target, causing domain bias and error propagation. 
Most recently, inspired by MAE~\cite{mae}, several concurrent works, \eg, M3AE~\cite{m3ae}, VLC~\cite{vlc} and MaskVLM~\cite{maskvlm}, transfer the pixel reconstruction task into VLP by simply adding low-level pixel reconstruction tasks on top of VLP models. Besides, some methods, \eg,  FLAVA~\cite{flava} and BEiT-3~\cite{beit3}, explore a different mid-level discrete visual token prediction task on vision modality. 
These VLP methods, however, neglect the semantic gap between low-level pixels (or mid-level visual tokens) and high-level multimodal representations, which can disturb the learning of semantically rich representations. 
Our MAMO unifies masked multimodal modeling of both vision and language as predicting the high-level latent representations and features/concepts, thus avoiding the potential negative impacts brought by the semantic gaps.

\subsection{Self-Distillation}
Self-distillation methods \cite{byol,dino} attempt to utilize history knowledge to drive the model learning from itself. 
BYOL~\cite{byol} proposes a self-supervised image representation learning approach that iteratively bootstraps the outputs of a network to serve as targets for enhanced representations.
Some methods~\cite{ibot, data2vec, cmae} combine this learning strategy with masked modeling for separate modalities.  
ALBEF~\cite{albef} use momentum-distillation strategy to generate pseudo logits, which helps model to improve learning from noisy web data.
The implicit prediction target of MAMO gets inspiration from these methods, but our work differs in that we operate on multimodal inputs and representations with the help of mask-and-predict tasks, and further enhance the representations with the aid of explicit and semantically meaningful targets.

\section{Method}

\subsection{Model Architecture}
\begin{figure}[t]
\centering
\includegraphics[width=0.8\linewidth]{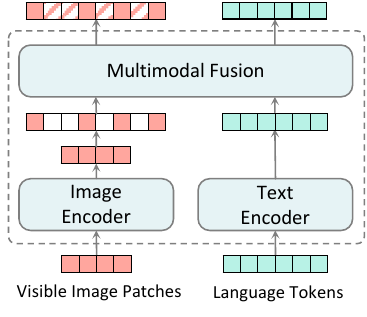}
\caption{Architecture of MAMO.}
\label{fig:detail_model}
\end{figure}

As illustrated in \cref{fig:detail_model}, our model includes an image encoder, a text encoder, and a multimodal fusion encoder. 
The image encoder is a pre-trained visual transformer, ViT-B/16~\cite{vit}. 
The text encoder and the multimodal fusion encoder are both based on transformer, which are initialized with the first 6 layers and the last 6 layers of BERT$_{base}$ \cite{bert}, respectively. 
The image encoder encodes an input image $I$ into a sequence of visual features $\{v_{cls},v_1,v_2,...,v_N \} $, where $v_{cls}$ represents the global feature on the [CLS] token and others correspond to each visible image patch. Particularly, we append a shared, learnable vector (\ie , mask token) into those visual features to indicate the presence of a missing patch to be predicted, and add positional embeddings to all tokens in this full set. 
The text encoder transforms an input text $T$ into a sequence of linguistic token features  $\{w_{cls},w_1,w_2,...,w_N \} $. 
The visual and linguistic tokens are concatenated and fed to the multimodal fusion encoder to generate fused multimodal representations. 

We pre-train our model with two categories of pre-training tasks: 1) masked multimodal modeling, which enables learning fine-grained multimodal interaction by the way of mask-and-predict; 2) global-level image-text alignment, which aligns image and text from the perspective of their global consistency.

\begin{figure*}[t]
\begin{center}
 \includegraphics[width=\linewidth]{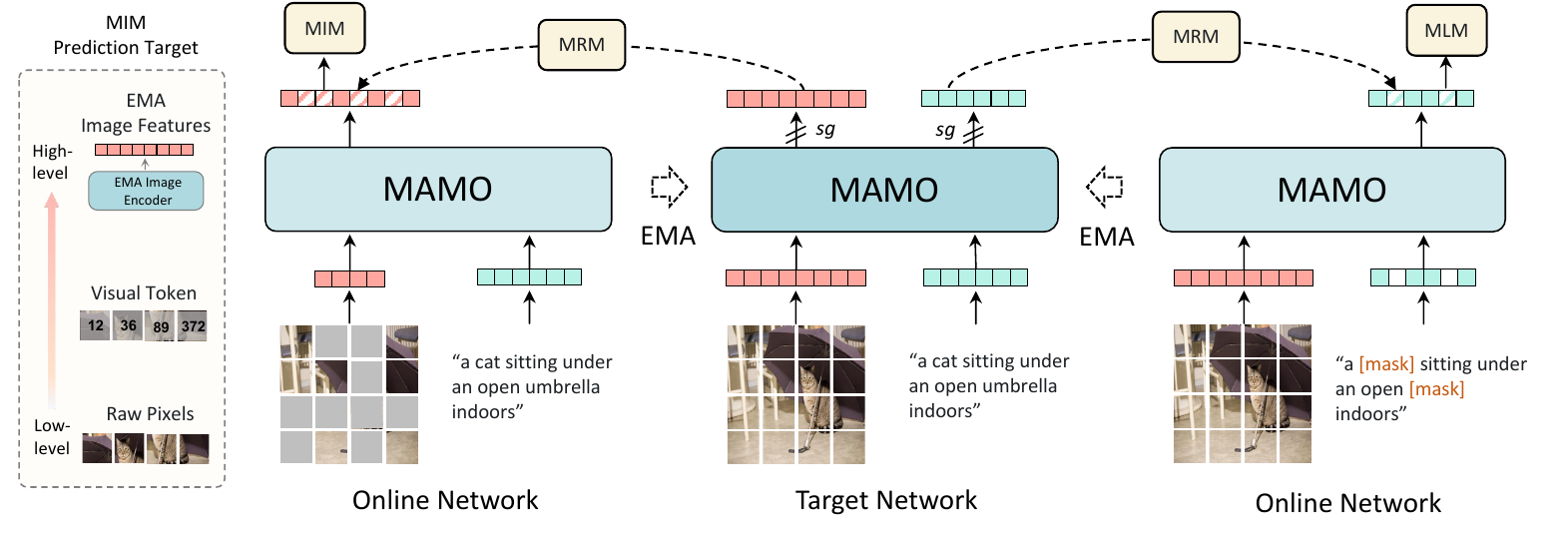}
\end{center}
\caption{Illustration of our masked multimodal modeling method. We enforce fine-grained multimodal interaction by jointly performing mask-and-predict tasks on both vision and language. MRM is an implicit, unified prediction target for vision and language that performs self-distillation between the online network and an exponentially moving averaged (EMA) target network. MIM and MLM are explicit, semantically meaningful prediction targets for image and text respectively. \textit{sg} means stop gradient.} 
\label{mamo}
\end{figure*}

\subsection{Masked Multimodal Modeling}

Masked multimodal modeling is designed to learn fine-grained multimodal interaction for vision-language input.
Given an image-text pair $(I,T)$, we create two masked views, $(\hat{I},T)$ and $(I,\hat{T})$, by randomly masking a portion of the input, \ie , either removing some image patches  or replacing some sub-words in the text with [MASK] token. 
The masked views, $(\hat{I},T)$ and $(I,\hat{T})$, are sent into our model, the online network $f$ parameterized by $\theta$, to get their multimodal representations, $f_\theta(\hat{I},T)$ and $f_\theta(I,\hat{T})$. We then design an implicit (masked representation modeling) and two explicit (masked image/language modeling) prediction sub-tasks for those masked views.

\paragraph{Masked Representation Modeling (MRM)} MRM serves as an implicit, unified, and debiased prediction target for both vision and language.
It requires predicting latent multimodal representations on each masked position under the guidance of a self-distillation teacher, which is referred to as the target network. 

The target network has the same architecture as the online network, both defined as $f$, but uses a different set of parameters. 
We don't simply copy the online network to the target network because the frequent change in the target network makes the learning process diverge. To acquire a smoothing target, we utilize a momentum target network~\cite{byol} whose parameters $\bar{\theta}$ are updated by an exponential moving average (EMA) of the online parameters $\theta$: $\bar{\theta} \gets \alpha  \bar{\theta} + (1-\alpha)  \theta$. 
We stop the gradient propagation in the target network.

We send the unmasked raw input ${(I, T)}$ into the target network to get its latent multimodal representation $f_{\bar{\theta}}(I,T)$. This full, unmasked representation then serves as the prediction target of $f_\theta(\hat{I},T)$ and $f_\theta(I,\hat{T})$ on masked positions. 
Following \cite{simclr}, we add a non-linear projector $g$ on both the online and target networks, which have been empirically shown beneficial to performance according to \cite{simclr}. We here define the MRM objective as:
\begin{equation}
\displaystyle
\begin{aligned}
\mathcal{L}_{mrm} = \mathbb{E}_{(I,T)\sim D}\bigg[\ell_{mrm}\left( f_\theta(I,\hat{T}), f_{\bar{\theta}}(I,T)\right) \\
+ \ell_{mrm}\left( f_\theta(\hat{I},T),  f_{\bar{\theta}}(I,T)\right) \bigg]
\end{aligned}
\end{equation}
The objective function $ \ell_{mrm}(x, y)$ calculates mean squared error (MSE) between predictions $ h_{mrm} \circ g_\theta(x)$ and targets $ g_{\bar{\theta}}(y)$ on each masked position. The projector $g$ and the predictor $h_{mrm}$ are both MLPs with one hidden layer.

MRM unifies the training objective of both vision and language, and avoids learning biases in modality-specific designs. 
Our momentum representation targets can be seen as a dynamic ensemble of historical targets, which offers varying target views that encourage the online network learning more general semantic representations. 

While MRM is effective empirically, such self-distillation may collapse into trivial solutions, \eg, outputting the same vector for all masked positions. To overcome such problems and further enrich the multimodal representations, we then introduce two explicit and semantically meaningful prediction targets for vision and language, respectively.

\paragraph{Masked Image Modeling (MIM)} MIM feeds $(\hat{I}, T)$ into the network, and predicts the masked image information with the help of the text and the visible image patches. 
In self-supervised learning of images, popular MIM targets include raw pixels~\cite{mae,simmim}, visual tokens~\cite{beit} from a pre-trained dVAE~\cite{dalle}, \etc, as are displayed in the left side of~\cref{mamo}.
However, pixels are low-level raw signals containing high-frequency details, while 
visual tokens~\cite{beit} are mid-level information that encapsulate mostly patch details (\eg color and texture, according to~\cite{ibot}).
Such low- and mid-level prediction targets have semantic gaps with high-level representations. Learning of such targets makes the model struggle in separating semantics and visual details, yielding less competitive performance in semantic understanding~\cite{liu2021self}. Also, the visual tokens rely on an offline pre-trained dVAE \cite{dalle} model, which may result in error propagation. 

In replacement of these previous image prediction targets, we introduce the momentum visual features, extracted by the target network's image encoder $f^v_{\bar{\theta}}$, to be a more appropriate prediction target of MIM in VLP. 
The momentum visual features are high-level and semantically meaningful, thus can avoid the aforementioned semantic gap. 
Our MIM objective can be formulated as:
\begin{equation}
    \mathcal{L}_{mim} = \mathbb{E}_{(I,T)\sim D}\ell_{mim}\left(f_\theta(\hat{I},T), f^v_{\bar{\theta}}(I,T)\right) 
\end{equation}
The objective function $\ell_{mim}(x,y)$ calculates L1 loss between predictions $h_{mim}(x)$ and targets $y$ on each masked image position, where $h_{mim}$ denotes an MLP predictor.

\paragraph{Masked Language Modeling (MLM)} MLM feeds $(I,\hat{T})$ into the network, and predicts the masked text information. Since word concepts in the text are already information-dense and semantically rich, we simply use word tokens as the explicit target, which is as same as the masked language modeling in~\cite{bert}. MLM predicts original word tokens on each masked text position.
The MLM objective is defined as:
\begin{equation}
    \mathcal{L}_{mlm}=\mathbb{E}_{(I,T)\sim D}\ell_{mlm}\left(f_\theta(I,\hat{T}), T\right) 
\end{equation}
The objective function $\ell_{mlm}(x,y)$ calculates cross-entropy loss between word prediction probability $h_{mlm}(x)$ and the ground-truth $y$ on each masked text position, where $h_{mlm}$ denotes a linear-softmax predictor.

\subsection{Global-level Image-Text Alignment}
Global-level image-text alignment is designed to learn global consistency between image and text from perspectives of unimodal and multimodal representations, respectively. These global-level alignment tasks have been widely used in previous VLP methods~\cite{albef, meter, vlmo}.

\paragraph{Image-Text Contrastive Learning (ITC)}
ITC~\cite{albef} facilitates the global alignment between the features of images and their corresponding texts before feeding them into the multimodal fusion encoder. 
Given a batch of image-text pairs $\{(I_j,T_j) \}_{j=1}^N$, ITC minimizes the InfoNCE~\cite{cpc} loss as: 
\begin{equation}
\begin{aligned}
    \mathcal{L}_{itc}=-\frac{1}{2}\mathbb{E}_{(I,T)\sim D}\bigg[log\frac{\exp(s(I,T)/\tau)}{\sum_{j=1}^N \exp(s(I,T_j)/\tau)} \\
    +  log\frac{\exp(s(I,T)/\tau)}{\sum_{j=1}^N \exp(s(I_j,T)/\tau)}\bigg]
\end{aligned}
\end{equation}
where $s(I,T)=g^v(v_{cls})^{\mathsf{T}}g^t(w_{cls})$ is the similarity function, $g^v$ and $g^t$ denote two projection heads and $\tau$ is a learnable temperature parameter~\cite{clip}.

\paragraph{Image-Text Matching Learning (ITM)} ITM aims to determine whether an image and a text are matched. We use the similarity probability in ITC to sample an in-batch hard negative example for each image and each text respectively. And then, we use the [CLS] token from multimodal fusion encoder's output to predict whether the given image-text pair is matched. The ITM loss is defined as:
\begin{equation}
    \mathcal{L}_{itm}=\mathbb{E}_{(I,T)\sim D}\ell_{itm}\left(f_\theta(I,T), y(I,T)\right)
\end{equation}
 The objective function $\ell_{itm}(x,y)$ calculates the cross-entropy loss between the matching probability $p(x_{cls})$ and the matching label $y$ for each sample.

Finally, the full pre-training objective of MAMO is:
\begin{equation}
    \mathcal{L}=\mathcal{L}_{mrm}+\mathcal{L}_{mim}+\mathcal{L}_{mlm}+\mathcal{L}_{itc}+\mathcal{L}_{itm}
\end{equation}

\section{Experiments}

\subsection{Pre-training Datasets}
Following previous works~\cite{uniter,albef}, we construct our image-text pre-training data using the union of two web datasets, \ie , Conceptual Captions~\cite{cc3m} and SBU Captions~\cite{sbu}, and two in-domain datasets, \ie , Visual Genome~\cite{vg} and MSCOCO Captions~\cite{coco}. 
The total number of unique images is 4.1M, and the number of image-text pairs is 9.7M. \Cref{table-data} shows the statistics of the pre-training datasets.

\subsection{Implementation Details}
Our model consists of a BERT$_{base}$~\cite{bert} with 110.1M parameters and a ViT-B/16~\cite{vit} with 85.9M parameters. We pre-train the model for 40 epochs using a batch size of 2048 on 32 NVIDIA A100 GPUs. We use the
AdamW~\cite{adamw}  optimizer with a weight decay of 0.01. The learning rate is warmed-up to $1\times10^{-4}$ in the first 2500 iterations, and decayed to $1\times10^{-5}$ following a linear schedule. During pre-training, we take
random image crops of resolution $224\times224$ as input, and also apply RandAugment~\cite{randaugment}.
We keep $25\%$ masking ratio on text modality, the replacements are $10\%$ random tokens, $10\%$ unchanged, and $80\%$ [MASK] token. And we apply random masks on image patches with a masking ratio of $75\%$.
Following previous works~\cite{albef,tcl,codis}, during fine-tuning, we increase the image resolution to $384\times384$ and interpolate the positional encodings of image patches following~\cite{vit}. The momentum parameter $\alpha$ for updating the target network is set as 0.995.

\begin{table}[t]
\centering
\caption{Statistics of the pre-training datasets.}
\label{table-data}
\begin{tabular}{c|cccc}
\toprule
 & MSCOCO & VG & SBU & CC-3M \\
 \midrule
 \#image & 113K & 100K & 860K & 2.95M \\
 \#text & 567K & 5.4M & 860K & 2.95M \\
\bottomrule
\end{tabular}

\end{table}

\subsection{Downstream Tasks}
We mainly apply our pre-trained model to four common V+L downstream tasks. More details are shown in \cref{details}. 

\textbf{Image-Text Retrieval} includes two sub-tasks: text-to-image retrieval (IR) and image-to-text retrieval (TR). We evaluate the performance on Flickr30K~\cite{flickr} and MSCOCO~\cite{coco} benchmarks on the fine-tuned and zero-shot settings. Following previous methods~\cite{albef,tcl,codis}, we use the model fine-tuned on MSCOCO for zero-shot evaluation on Flickr30K. During fine-tuning, we trained model with ITC and ITM loss.  During inference, we select the top-k (256 for MSCOCO, and 128 for Flickr30K) candidates using ITC scores and
calculate their ITM scores to rank these candidates following~\cite{albef}, which accelerates the inference speed and keeps the retrieval performance.

\begin{table*}[t]
  \caption{Fine-tuned image-text retrieval results on MSCOCO and Flickr30K datasets. IR: image retrieval. TR: text retrieval.}
  \label{table-ret}
 \centering
 \resizebox{\linewidth}{!}{
\begin{tabular}{ll|cc|cc}
\toprule
\multirow{3}{*}{Method} & \multirow{3}{*}{\#Images} & \multicolumn{2}{c|}{MSCOCO (5k test set)}                                     & \multicolumn{2}{c}{Flickr30K (1k test set)}                                  \\
                        &                                  & TR                                   & IR                                    & TR                                   & IR                                   \\
\multicolumn{1}{l}{}    & \multicolumn{1}{l|}{}            & \multicolumn{1}{l}{R@1 / R@5 / R@10} & \multicolumn{1}{l|}{R@1 / R@5 / R@10} & \multicolumn{1}{l}{R@1 / R@5 / R@10} & \multicolumn{1}{l}{R@1 / R@5 / R@10} \\
\midrule
UNITER$_{large}$~\cite{uniter}          & 4M                               & 65.7 / 88.6 / 93.8                   & 52.9 / 79.9 / 88.0                    & 87.3 / 98.0 / 99.2                   & 75.6 / 94.1 / 96.8                   \\
ALBEF ~\cite{albef}                 & 4M                               & 73.1 / 91.4 / 96.0                   & 56.8 / 81.5 / 89.2                    & 94.3 / 99.4 / 99.8                   & 82.8 / 96.7 / 98.4                   \\
TCL~\cite{tcl}& 4M & 75.6 / 92.8 / 96.7 & 59.0 / 83.2 / 89.9 & 94.9 / 99.5 / 99.8 & 84.0 / 96.7 / 98.5 \\
CODIS~\cite{codis} & 4M & 75.3 / 92.6 / 96.6 & 58.7 / 82.8 / 89.7 & 95.1 / 99.4 / 99.9 & 83.3 / 96.1 / 97.8 \\
VLC~\cite{vlc} & 5.6M & 71.3 / 91.2 / 95.8 & 50.7 / 78.9 / 88.0 & 89.2 / 99.2 / 99.8 & 72.4 / 93.4 / 96.5 \\
VLMo$_{Base}$~\cite{vlmo}    & 4M   & 74.8 / 93.1 / 96.9 & 57.2 / 82.6 / 89.8        & 92.3 / 99.4 / 99.9 & 79.3 / 95.7 / 97.8 \\
METER-Swin~\cite{meter}   & 4M          & 73.0 / 92.0 / 96.3 & 54.9 / 81.4 / 89.3 & 92.4 / 99.0 / 99.5 & 79.0 / 95.6 / 98.0 \\
MaskVLM~\cite{maskvlm}   & 4M              & 76.3 / 93.8 / 96.8 & 60.1 / 83.6 / 90.4   & 95.6 / 99.4 / 99.9 & 84.5 / 96.7 / 98.2     \\
\midrule

ALIGN    ~\cite{align}              & 1.8B                             & 77.0 / 93.5 / 96.9                   & 59.9 / 83.3 / 89.8                    & 95.3 / \textbf{99.8} / \textbf{100.0}                  & 84.9 / \textbf{97.4} / \textbf{98.6}    
\\ \midrule
MAMO                  & 4M                               & \textbf{79.1} / \textbf{94.9} / \textbf{97.8}                  & \textbf{62.4} / \textbf{85.3} / \textbf{91.3}                    & \textbf{96.2} / 99.5 / 99.8                     & \textbf{86.1} / 97.0 / 98.4                   \\ 
\bottomrule
\end{tabular}}

\end{table*}

\begin{table}[t]
        \caption{Zero-shot image-text retrieval results on Flickr30K. 
        }
  \label{table-ret-zs}
    \centering
\resizebox{\linewidth}{!}{
\begin{tabular}{ll|cc}
\toprule
\multirow{3}{*}{Method} & \multirow{3}{*}{\#Images} &  \multicolumn{2}{c}{Flickr30K (1k test set)}                                  \\
                        &                                                                    & TR                                   & IR                                   \\
\multicolumn{1}{l}{}    & \multicolumn{1}{l|}{}            & \multicolumn{1}{l}{R@1 / R@5 / R@10} & \multicolumn{1}{l}{R@1 / R@5 / R@10}  \\
\midrule
UNITER$_{large}$~\cite{uniter}          & 4M             & 83.6 / 95.7 / 97.7                  & 68.7 / 89.2 / 93.9                                 \\
ALBEF ~\cite{albef}                 & 4M                               & 90.5 / 98.8 / 99.7                   & 76.8 / 93.7 / 96.7             \\
\midrule
CLIP~\cite{clip} & 400M & 88.0 / 98.7 / 99.4 & 68.7 / 90.6 / 95.2  \\
ALIGN~\cite{align} & 1.8B & 88.6 / 98.7 / 99.7 & 75.7 / 93.8 / 96.8  \\
\midrule

MAMO                  & 4M                               & \textbf{93.0} / \textbf{99.5} / \textbf{99.9}                   & \textbf{80.6} / \textbf{95.2} / \textbf{97.2}                                 \\ 
\bottomrule
\end{tabular}
}

\end{table}

\textbf{Visual Question Answering (VQA)} requires the model to predict an answer given an image and a question. We evaluate our model on VQA2.0 dataset~\cite{vqa}. Following previous methods~\cite{VL-T5, albef}, we first fuse the question and image with our pre-trained model, and then input the multimodal representation sequence to a 6-layer transformer decoder to auto-regressively generate answers. %

\textbf{Natural Language for Visual Reasoning (NLVR)} lets the model determine whether a given text describes the relationship between two images. We evaluate our model on NLVR2 dataset~\cite{nlvr}. Following~\cite{uniter}, we create two image-text pairs using the same text and different images, and get their respective multimodal representations. And then we use bidirectional attention to establish interactions between the two outputs. %

\textbf{Visual Grounding} aims to localize the corresponding region in an image with a specific text description. Following~\cite{albef}, we study weakly-supervised visual grounding without any available bounding box annotations on the RefCOCO+~\cite{refcoco} dataset. We fine-tune our model with global-level image-text alignment tasks (\ie , ITC and ITM) using only image-text supervision. During inference, we utilize Grad-CAM~\cite{gradcam} to get image heatmaps and use them to rank the region proposals provided by~\cite{mattnet}. 
\subsection{Results on V+L downstream tasks}

\begin{table}[t]
        \caption{Comparison with state-of-the-art methods on VQA and NLVR2.}
  \label{table-vr}
    \centering
\resizebox{0.9\linewidth}{!}{
\begin{tabular}{lll|cc}
\toprule
\multirow{2}{*}{Method} & \multicolumn{2}{c|}{VQA}                & \multicolumn{2}{c}{NLVR2}                \\
                        & test-dev           & test-std           & dev                 & test-P             \\ \midrule
LXMERT   ~\cite{lxmert}              & 72.42              & 72.54              & 74.90               & 74.50              \\
UNITER$_{large}$~\cite{uniter}           & 73.82 & 74.02 & 79.12 & 79.98 \\
OSCAR   ~\cite{oscar}               & 73.16              & 73.44              & 78.07               & 78.36              \\
ALBEF ~\cite{albef}             & 74.54              & 74.70              & 80.24               & 80.50              \\
TCL     ~\cite{tcl}          & 74.90              & 74.92              & 80.54               & 81.33             \\
CODIS ~\cite{codis}& 74.86 & 74.97 & 80.50 & 80.84 \\
VLC (5.6M)~\cite{vlc} & 74.02 & 74.00 & 77.70 & 79.04 \\
VLMo$_{Base}$~\cite{vlmo}              & 76.64      & 76.89      & 82.77       & 83.34       \\
METER-Swin   ~\cite{meter}             & 76.43      & 76.42      & 82.23       & 82.47       \\
MaskVLM    ~\cite{maskvlm}             & 75.45      & 75.40      & 81.58       & 81.98       \\
\midrule
MAMO                   & \textbf{77.07}              &\textbf{77.16}              & \textbf{83.58}
& \textbf{84.18}
\\
\bottomrule
\end{tabular}
}

\end{table}

\paragraph{Results on Image-Text Retrieval}
\Cref{table-ret,table-ret-zs} report the results on fine-tuned and zero-shot image-text retrieval respectively. Our method achieves state-of-the-art (SoTA) performance compared with other methods on all retrieval tasks. On the zero-shot retrieval task, our method outperforms ALBEF~\cite{albef} by 3.8\% on image retrieval R@1 and 2.5\% on text retrieval R@1. Although trained with only 4M data, our method even outperforms ALIGN~\cite{align} which is trained on a larger amount of 1.8B images. On the fine-tuned setting, we also achieve comparable performance with ALIGN~\cite{align} and outperform most previous works trained on a similar amount of data. 
These superior performances in both 
zero-shot and fine-tuned retrieval tasks validate that MAMO can generate semantically rich multimodal representations.

\paragraph{Results on VQA and NLVR}
As is shown in~\cref{table-vr}, our method achieves SoTA performance both on VQA~\cite{vqa} and NLVR2~\cite{nlvr}. Though our method doesn't use any external object annotations during pre-training, it still outperforms most region-based methods~\cite{uniter,oscar} that benefit from object structures in the regional features. Compared to ALBEF~\cite{albef}, we achieve improvements of 2.46\% on VQA test-std and 3.68\% on NLVR2 test-P.

\paragraph{Results on Weakly-Supervised Visual Grounding}
\Cref{table-vg} reports the results on weakly-supervised visual grounding. Our method outperforms previous weakly-supervised method~\cite{albef} and the baseline model pre-trained without masked multimodal modeling. The results suggest that our method helps the model locate semantic regions in images and learn more fine-grained interaction between words and image regions.

\subsection{Ablation Study}
For ablation studies, we train all models with a shorter training scheduler of 10 epochs. We report R@1 on fine-tuned retrieval tasks and accuracy on VQA.

\begin{table}[t]
  \captionof{table}{Weakly-supervised visual grounding results on RefCOCO+.}
  \label{table-vg}
\centering
 \resizebox{0.8\linewidth}{!}{
\begin{tabular}{l|ccc}
\toprule

Method                           &   val      & testA & testB \\
                               \midrule
ALBEF \cite{albef}                        & 58.46         & 65.89 & 46.25 \\
MAMO w/o mask                    & 56.61        & 65.08 & 42.91\\
MAMO                    & \textbf{61.90}         & \textbf{72.30} & \textbf{48.50}\\
\bottomrule
\end{tabular}}

\end{table}

\begin{table}[t]
  \caption{Ablation experiments on pre-training tasks. IR: image retrieval recall@1. TR: text retrieval recall@1. }
  \label{table-ptt}
\centering
 \resizebox{\linewidth}{!}{
\begin{tabular}{cccc|cc|cc|c}
\toprule
\multicolumn{4}{c|}{Tasks}           &
 \multicolumn{2}{c|}{Flickr30K}                & \multicolumn{2}{c|}{MSCOCO}    & VQA            \\
                            ITM+ITC   & MRM & MLM & MIM  & TR & IR  & TR & IR & test-dev  \\
                               \midrule
   \checkmark   &  &  &                & 93.5 & 80.7   & 71.9  & 55.2  & 72.1 \\
 \checkmark   &  & \checkmark  &                    & 93.2   & 83.7    & 75.5  & 58.0 & 73.9   \\
  \checkmark   &  & \checkmark  &    \checkmark         & 94.0   & 83.5  & 74.7 &  58.2 & 74.4 \\
 \checkmark   & \checkmark &  &        & 94.1  & 81.4    & 71.8  & 55.5 & 72.9  \\
 \checkmark   & \checkmark & \checkmark &         & 95.2 & 83.6   & 75.8  & 59.0 & 74.6 \\

 \checkmark   & \checkmark   & \checkmark   &    \checkmark             & \textbf{95.5} & \textbf{84.3}   & \textbf{76.6} & \textbf{59.2} & \textbf{74.8} \\
\bottomrule
\end{tabular}}

\end{table}

\begin{table}
  \captionof{table}{Ablation experiments on MRM modalities. }
  \label{table-mrmmod}
		\centering
 \resizebox{0.8\linewidth}{!}{
\begin{tabular}{cc|cc|cc|c}
\toprule
\multicolumn{2}{c|}{MRM modality}               &
 \multicolumn{2}{c|}{Flickr30K}                & \multicolumn{2}{c|}{MSCOCO}    & VQA            \\
                      image & text        & TR & IR  & TR & IR & test-dev  \\
                              \midrule
 &            & 94.0 & 83.5   & 74.7  & 58.2  & 74.4 \\
  \checkmark &                 & 93.6   & 84.4    & 75.2  & 58.8 & 74.7   \\
  & \checkmark                 & 94.7   & 84.2    & 75.6  & 58.8 & 74.7   \\
  \checkmark & \checkmark              & \textbf{95.5} & \textbf{84.3}   & \textbf{76.6} & \textbf{59.2} & \textbf{74.8} \\
\bottomrule
\end{tabular}
}

\end{table}

\paragraph{Importance of Pre-training Tasks}
We evaluate the effectiveness of our pre-training tasks on \cref{table-ptt}. 
Adding our implicit target MRM on all other tasks brings improvements to almost all downstream tasks. But using MRM alone only slightly improves the performance, which we infer is because the implicit target may collapse into trivial solutions that restrict the modeling capacity, \eg, producing the same values on some feature dimensions.
When we combine the implicit MRM with the auxiliary explicit targets, MIM and MLM, the model achieves significantly better performances on both VQA and retrieval tasks, and is also better than using only explicit MLM and MIM targets, suggesting that our implicit and explicit prediction targets are highly complementary.

\paragraph{Influence of MRM Modalities}
We explore the separate effects of MRM on vision and language modalities. We compare 4 different MRM settings: 1) without MRM, 2) MRM on image patches only, 3) MRM on text tokens only, and 4) MRM on both image and text. As is shown in \cref{table-mrmmod}, MRM is effective on both modalities. Meanwhile, applying MRM to both modalities simultaneously brings larger performance improvement.  

\paragraph{Design Choices of MRM}
In \cref{table-mrmdesign}, we perform ablation studies on the design details of MRM. When we replace the EMA updating target network with the online network on the last training step, we find MRM loss diverges, resulting in terrible performance on downstream tasks. We argue that the frequent change of prediction targets harms the training process. On the contrary, the EMA updating target network smooths down the prediction targets, leading to a stable learning process. When we remove the asymmetric predictor in MRM, there is a performance degradation on all downstream tasks, which indicates the necessity of the predictor since the online representations and target representations are not in the same semantic space.

\begin{table}[]
  \caption{Ablation experiments on MRM design. $\times$ mean the model diverges during pre-training, leading to terrible performance on downstream tasks.}
  \label{table-mrmdesign}
\centering

 \resizebox{0.85\linewidth}{!}{
\begin{tabular}{l|cc|cc|c}
\toprule
\multicolumn{1}{l|}{\multirow{2}{*}{MRM design}}               &
 \multicolumn{2}{c|}{Flickr30K}                & \multicolumn{2}{c|}{MSCOCO}    & VQA            \\
                                & TR & IR  & TR & IR & test-dev  \\
                               \midrule
   MAMO   &       \textbf{95.5} & \textbf{84.3}   & \textbf{76.6} & \textbf{59.2} & \textbf{74.8} \\
  w/o EMA update           & $\times$ & $\times$   & $\times$  & $\times$  & $\times$ \\
  w/o predictor                & 94.1   & 83.8    & 75.6  & 58.4 & 74.4   \\

\bottomrule
\end{tabular}
}

\end{table}
\begin{table}[t]
        \caption{Ablation experiments on MIM prediction targets. We grid search the best decoder depth and report the best results for each variant.}
        \label{table-mim-target}
    \centering
 \resizebox{0.95\linewidth}{!}{
\begin{tabular}{l|cc|cc|c}
\toprule
\multirow{2}{*}{MIM target}           &
 \multicolumn{2}{c|}{Flickr30K}                & \multicolumn{2}{c|}{MSCOCO}    & VQA            \\
                              & TR & IR  & TR & IR & test-dev  \\
                              \midrule
 raw pixels            & 93.1 & 83.9   & 75.8  & 59.1  & 74.5 \\
 visual tokens                 & 95.2   & 84.0    & 76.5  & 58.1 & 74.6   \\

  momentum visual features             &   \textbf{95.5} & \textbf{84.3}   & \textbf{76.6} & \textbf{59.2} & \textbf{74.8} \\
\bottomrule
\end{tabular}}

\end{table}

\begin{table}[t]       
\caption{Ablation experiments on MIM decoder depth. 0 refers to a simple MLP projector.}
        \label{table-mim-dec}
    \centering
 \resizebox{0.7\linewidth}{!}{
\begin{tabular}{c|cc|cc|c}
\toprule

    \multirow{2}{*}{\#Blocks}      &
 \multicolumn{2}{c|}{Flickr30K}                & \multicolumn{2}{c|}{MSCOCO}    & VQA            \\
                         & TR & IR  & TR & IR & test-dev  \\
                              \midrule
0              & \textbf{95.5} & 84.3   & \textbf{76.6} & 59.2 & \textbf{74.8} \\
2                & 94.7   & 84.1    & 75.9  & \textbf{59.3} & \textbf{74.8}   \\
  4               & 94.7   & \textbf{84.6}    & 76.1  & 59.0 & 74.7   \\
6            & 95.0 & 84.2   & 75.8  & \textbf{59.3}  & 74.7 \\

\bottomrule
\end{tabular}}

\end{table}

\paragraph{MIM Prediction Targets}
In \cref{table-mim-target}, we compare different MIM prediction targets including raw pixels, visual tokens, and momentum visual features. 
We grid search the best decoder depth for each variant.
For the raw pixels target, we use a 4-layer transformer decoder. For visual tokens, we use an MLP predictor to predict visual tokens extracted by pre-trained DALL-E~\cite{dalle} codebook. For momentum visual features, we also use an MLP predictor as the decoder. We can see that the high-level momentum visual features target outperforms other low- and mid-level prediction targets.

\paragraph{Decoder Depth of MIM}
Some masked image modeling methods such as MAE~\cite{mae} adopt an asymmetric encoder-decoder architecture. When it comes to MIM with momentum visual features as the target, we also explore whether a decoder is vital to help the model extract better representations. As shown in \cref{table-mim-dec}, we use either transformer or MLP decoder and compare different decoder depths. We find that decoder depth has little influence on our performance. Among all these architectures, an MLP layer (depth 0) performs well with the simplest design.

\begin{figure*}[t]
    \centering

	\begin{minipage}{0.49\linewidth}
		\centering
		\includegraphics[width=0.8\linewidth]{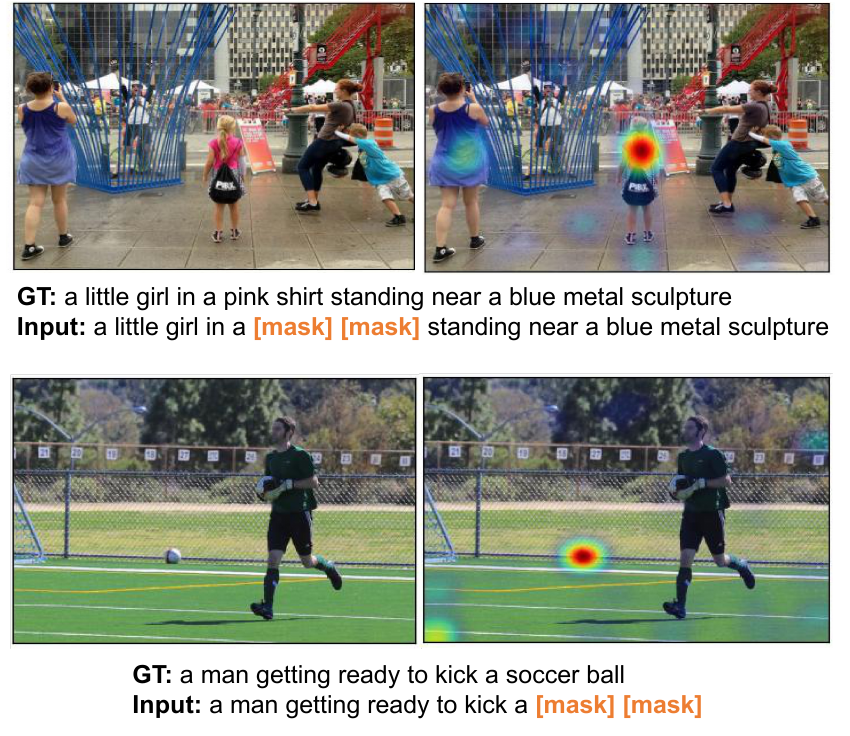}
		\caption{Grad-CAM visualization of MRM on masked text tokens. }
		\label{fig:1}
	\end{minipage}
	\hfill
	\begin{minipage}{0.45\linewidth}
		\centering
		\includegraphics[width=0.85\linewidth]{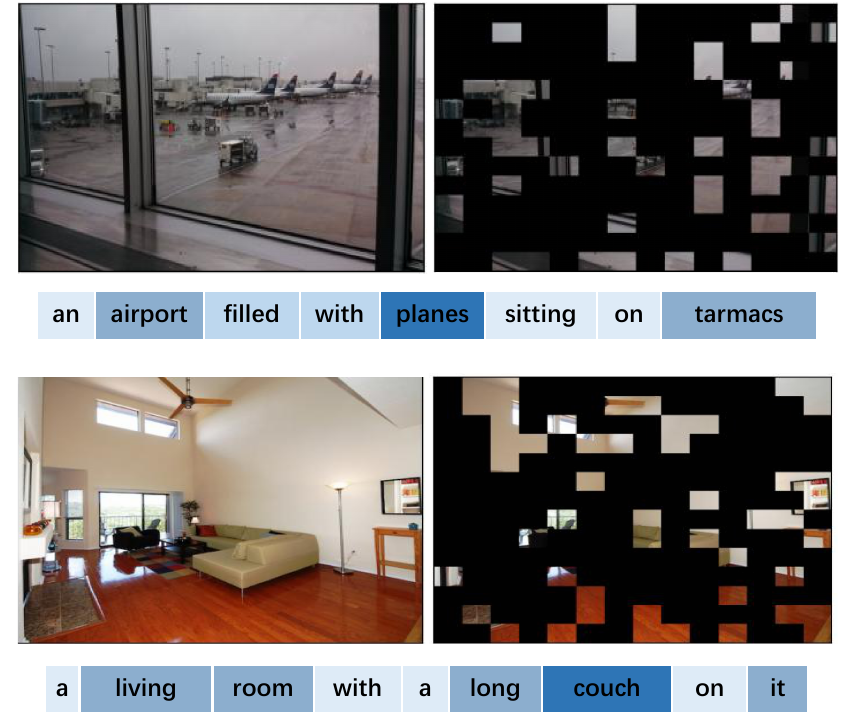}
		\caption{Grad-CAM visualization of MRM on masked image patches. A deeper color on the word means a higher Grad-CAM value.}
		\label{fig:2}
	\end{minipage}
\end{figure*}

\begin{figure}[t]	
		\centering
		\includegraphics[width=\linewidth]{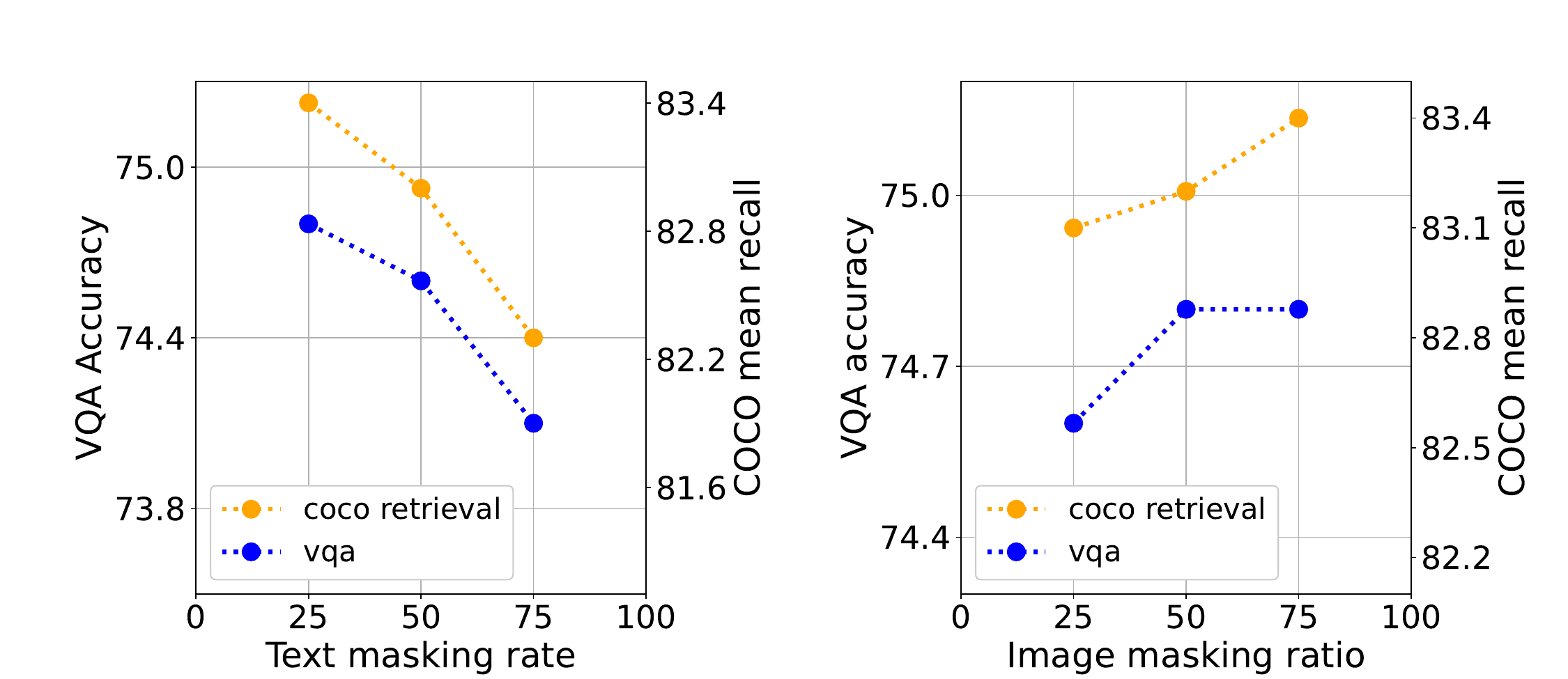}
		\caption{Choices of masking ratio for text and image. The accuracy on VQA test-dev set and the mean recall on MSCOCO retrieval are reported.}
		\label{fig-maskrate}
\end{figure}

\paragraph{Choices of Masking Ratio}
We compare different masking ratios for MAMO on both image and text modalities in \cref{fig-maskrate}. We find that a larger masking ratio on text brings performance degradation while a larger masking ratio on image generally performs better on V+L downstream tasks. We argue that a large masking ratio in the highly compressed text almost destroys the sentence structure. On the contrary, raw images are highly redundant natural signals, requiring a larger masking ratio. 

\subsection{Effect of Pre-Training Data Size}
\begin{figure}[t]
\vspace{-0.5cm}
		\centering
		\includegraphics[width=\linewidth]{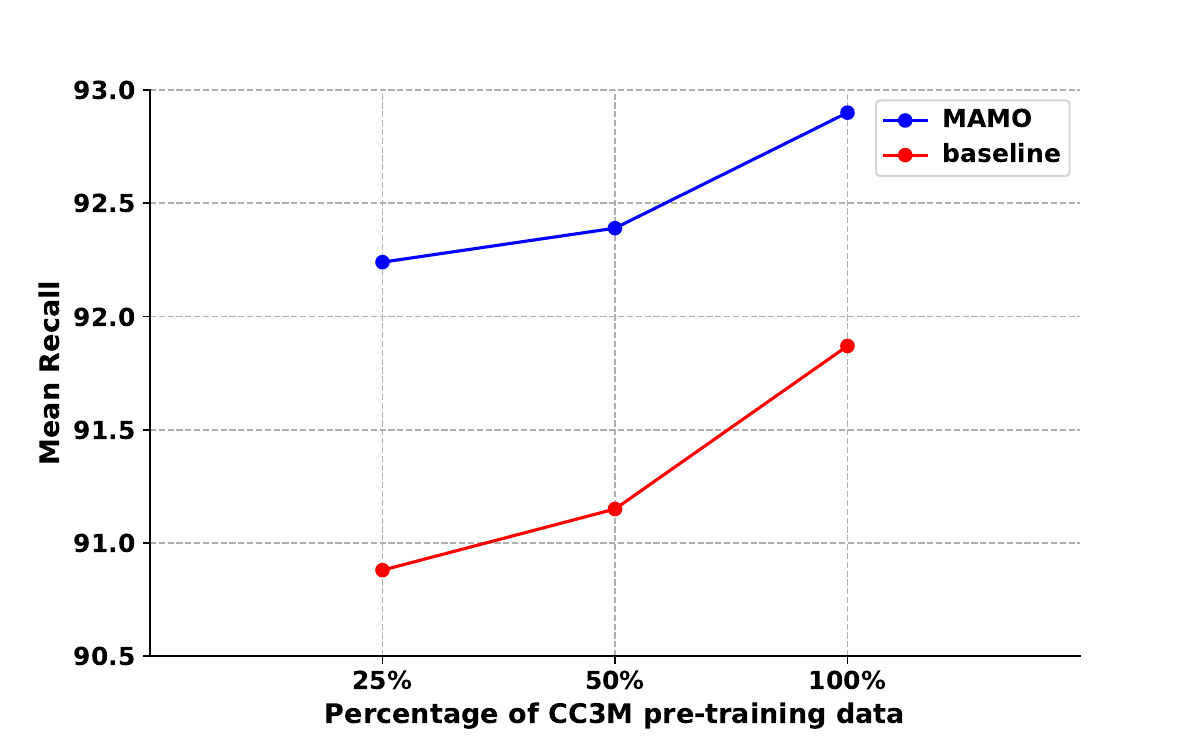}
		\caption{Effect of pre-training data size. The mean recall of zero-shot image-text retrieval on Flickr30K test set is reported. }
		\label{fig-partcc}
		\vspace{-0.5cm}
\end{figure}

We explore the effect of pre-training data size. We randomly sample $25\%$, $50\%$, and $100\%$ of CC-3M data and combine them with the other datasets in \cref{table-data} (\ie  , MSCOCO+VG+SBU). The number of image-text pairs in each subset is around $57\%$, $71\%$, and $100\%$ of the full 4M pre-training data, respectively. We show the result of our model MAMO and a baseline model pre-trained with ITM+ITC+MLM tasks in \cref{fig-partcc}. Mean recall in \cref{fig-partcc} is the average of image and text recall@1/@5/@10. Both models enjoy a performance improvement while the number of pre-training data increases. Meanwhile, MAMO pre-trained with only $25\%$ CC-3M data still outperforms the baseline model pre-trained with $100\%$ CC-3M data.

\subsection{Visualization}

We use Grad-CAM~\cite{gradcam} heatmap to showcase the self-attention maps between image and text tokens of the pre-trained model, where the gradients are acquired by minimizing the MRM loss of the masked tokens. 
We provide the Grad-CAM visualizations for MRM on both masked image patches (MRM-image) and masked text tokens (MRM-text) on the 4th layer of the multimodal fusion encoder. When studying MRM-text, we mask some words in the text and show the Grad-CAM heatmap on the corresponding image.  When studying MRM-image, we randomly mask some patches in the image and show the Grad-CAM values on the corresponding text. We provides a few visualizations in~\cref{fig:1,fig:2}. 
The visualization examples show that when MAMO predicts for masked tokens, it tends to put higher attention weights on the semantically correlated tokens in the other modality, indicating fine-grained cross-modal interaction is built during masked multimodal modeling learning.

\section{Conclusion}
In this work, we point out the shortcoming of existing VLP methods
in building fine-grained image-text interaction, which hurts performances on lots of V+L downstream tasks.
We tackle the problem by proposing  a jointly masked multimodal modeling method to learn fine-grained multimodal representations for VLP models. 
For efficient learning of jointly masked image-text signals, we integrate both implicit and explicit prediction targets that are both high-level and complementary to each other. 
Extensive experimental results on diverse downstream V+L tasks show that compared to existing methods, our method offers better performance on both zero-shot and fine-tuned settings.
We believe our joint masked multimodal modeling method can be further expanded to multimodal pretraining situation with more modalities.

\begin{acks}
This work was supported by the National Key Research and Development Program of China (No. 2020AAA0106400), National Natural Science Foundation of China (U21B2043) and CAAI-Huawei MindSpore Open Fund. 
\end{acks}


\appendix
\begin{figure*}[t]
\begin{center}
\includegraphics[width=0.9\linewidth]{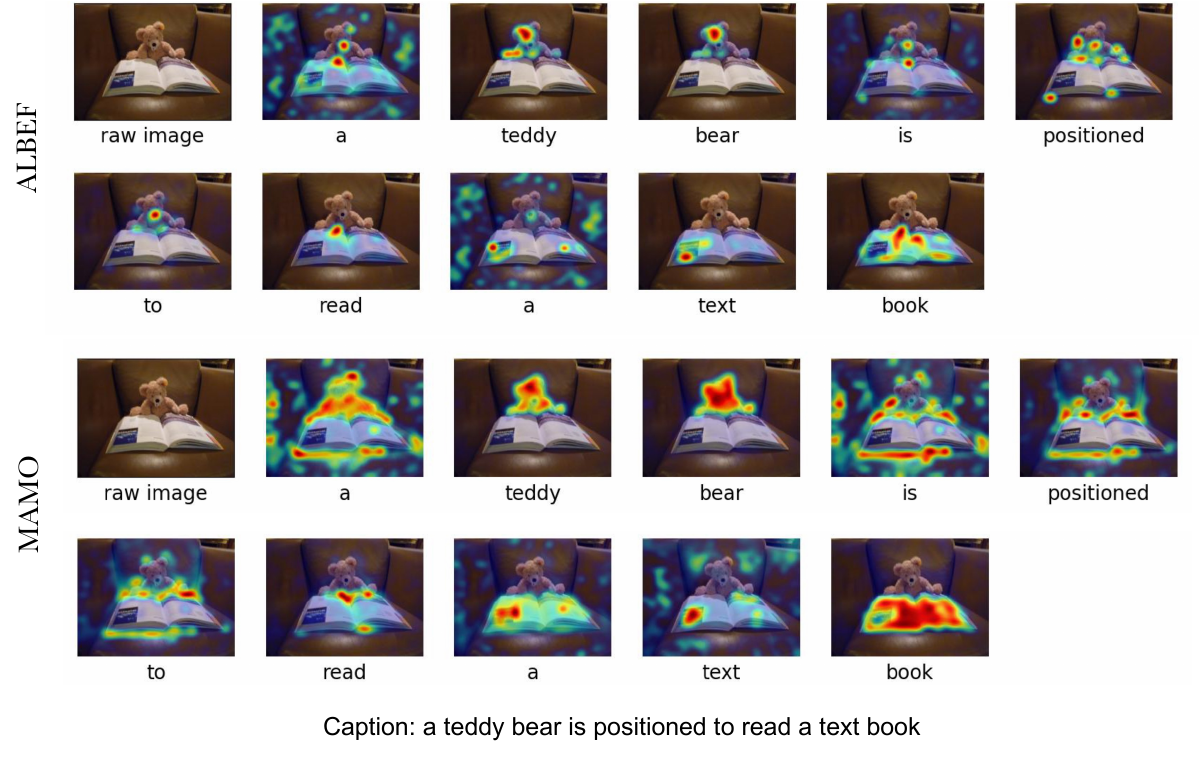}
\end{center}
\caption{Examples of per-word Grad-CAM visualization. MAMO can localize each word concept more precisely and comprehensively comparing with ALBEF~\cite{albef}. Instead of only focusing on some scattered small regions, MAMO more finely localizes the whole objects, showing a solid ability to separate different objects. }
\label{ex-perword1}
\end{figure*}

\section{Details on Fine-Tuning for Downstream Tasks}
\label{details}
For all downstream tasks, we use AdamW~\cite{adamw} optimizer with a weight decay of 0.01 and a linear learning rate scheduler. All downstream tasks take images with a $384\times 384$ resolution as inputs. 

\paragraph{Image-Text Retrieval} We fine-tune our model on two retrieval datasets: MSCOCO~\cite{coco} and Flickr30K~\cite{flickr}. We split the two datasets into train / val / test sets separately following the widely-used Karpathy split~\cite{karpathy}. We fine-tune our model for 10 epochs with a batch size of 160 and an initial learning rate of $3\times 10^{-5}$, optimizing the ITC and ITM loss. During inference, we select the top-$k$ (256 for MSCOCO, and 128 for Flickr30K) candidates using ITC scores and calculate their ITM scores to rank these candidates following \cite{albef}.
\paragraph{Visual Question Answering (VQA)} Following previous methods~\cite{albef}, we use both train and validation sets in VQA2.0 dataset~\cite{vqa} and the question-answer pairs from Visual Genome~\cite{vg} for training. Our VQA model uses a cross-attention decoder to generate answers. We initialize the decoder using the weights from our pre-trained multimodal fusion encoder and fine-tuning the VQA model for 10 epochs with a batch size of 192 and an initial learning rate of $5\times 10^{-5}$, optimizing the conditional language-modeling loss. To fairly compare with other methods, we constrain the decoder output to only generate 3,129 candidate answers during inference. Performance on the test-dev and test-std splits are reported.

\paragraph{Natural Language for Visual Reasoning (NLVR)} We evaluate our model on NLVR2 dataset~\cite{nlvr}. In NLVR task, the model receives one text and two input images. Since our model only handles one image and one text, we combine each image with the text to create two image-text pairs and extract their multimodal representations separately. Following ~\cite{uniter}, we use a bi-attention module to fuse the above two multimodal representations. We attach an FC layer to the concatenated bi-attention pair to get the result score for true-or-false classification.  We fine-tune the NLVR model for 10 epochs with a batch size of 80 and an initial learning rate of $3\times 10^{-5}$.

\paragraph{Visual Grounding} We fine-tune our weakly-supervised visual grounding model on RefCOCO+ dataset~\cite{refcoco}. We use global-level image-text alignment tasks (ITC and ITM) to train the model while removing random cropping. We fine-tune the model for 5 epochs with a batch size of 160 and an initial learning rate of $2\times 10^{-5}$. During inference, we use ITM loss to calculate Grad-CAM~\cite{gradcam} heatmap on the 4th self-attention layer in the multimodal fusion encoder. And we use the Grad-CAM value to select the best bounding box provided by~\cite{mattnet}.
\section{Visualization of Per-Word Localization}
In \cref{ex-perword1}, we show the Grad-CAM heatmap for each word in texts. It's worth noting that our MAMO can localize each word concept more precisely and comprehensively comparing with ALBEF~\cite{albef}. Instead of only focusing on some scattered small regions, MAMO more finely localizes the whole objects, showing a solid ability to separate different objects.

\bibliographystyle{ACM-Reference-Format}
\bibliography{main}

\end{document}